\documentclass{article} 

\usepackage[table]{xcolor}
\definecolor{linkColor}{rgb}{0.18,0.39,0.62}
\usepackage[preprint]{neurips_2021}

\usepackage{amsmath,amsfonts,bm}

\def\eqref#1{equation~\ref{#1}}

\def\1{\bm{1}}

\def\vb{{\bm{b}}}

\def\ve{{\bm{e}}}

\def\vh{{\bm{h}}}

\def\vo{{\bm{o}}}

\def\vt{{\bm{t}}}

\def\vv{{\bm{v}}}

\def\vx{{\bm{x}}}

\def\vz{{\bm{z}}}

\def\mS{{\bm{S}}}

\def\mW{{\bm{W}}}

\DeclareMathAlphabet{\mathsfit}{\encodingdefault}{\sfdefault}{m}{sl}
\SetMathAlphabet{\mathsfit}{bold}{\encodingdefault}{\sfdefault}{bx}{n}

\def\gD{{\mathcal{D}}}

\def\gM{{\mathcal{M}}}

\def\gV{{\mathcal{V}}}

\newcommand{\R}{\mathbb{R}}

\newcommand{\softmax}{\mathrm{softmax}}

\DeclareMathOperator*{\argmin}{arg\,min}

\usepackage[utf8]{inputenc} 
\usepackage[T1]{fontenc}    
\usepackage[colorlinks=true,linkcolor=linkColor,citecolor=linkColor,filecolor=linkColor,urlcolor=linkColor]{hyperref}       
\usepackage{url}            
\usepackage{booktabs}       
\usepackage{amsfonts}       
\usepackage{nicefrac}       
\usepackage{microtype}      
\usepackage{enumitem}
\usepackage{pifont}
\usepackage{xspace}
\usepackage{subfig}
\usepackage{multirow}

\RequirePackage{algorithm}
\RequirePackage{algorithmic}

\usepackage{multirow}
\usepackage{makecell}
\usepackage{amsmath}
\usepackage{capt-of}
\usepackage{tabularx}
\usepackage{epsfig}
\usepackage{amssymb}
\usepackage{amsfonts}
\usepackage{booktabs}
\usepackage{scalerel}
\usepackage{listings}
\usepackage{varwidth}
\usepackage[export]{adjustbox}
\usepackage{tikz}
\usetikzlibrary{tikzmark}

\usepackage{stmaryrd}
\usepackage{bbm}
\usepackage{wrapfig}
\usepackage{pifont}

\newcommand{\sptk}[1]{\texttt{[#1]}}

\definecolor{deepblue}{rgb}{0,0,0.5}
\definecolor{officeblue}{RGB}{0,102,204}
\definecolor{deepred}{rgb}{0.6,0,0}
\definecolor{deepgreen}{rgb}{0,0.5,0}
\definecolor{mybrickred}{RGB}{182,50,28}

\definecolor{fillcolor}{RGB}{216,217,252}

\definecolor{deemph}{gray}{0.6}
\newcommand{\gc}[1]{\textcolor{deemph}{#1}}
\definecolor{baselinecolor}{gray}{.9}
\newcommand{\baseline}[1]{\cellcolor{baselinecolor}{#1}}

\newcommand{\cmark}{\ding{51}\xspace}%
\newcommand{\xmark}{\ding{55}\xspace}%

\newcommand\beit{\textsc{BEiT}}
\newcommand\our{\textsc{BEiT v2}}
\newcommand\vqkd{\textsc{VQ-KD}}
\newcommand\vae{\textsc{VQ-VAE}}

\newcommand\cls{\sptk{CLS}}

\newcommand{\ie}{\textit{i}.\textit{e}.}
\newcommand{\eg}{\textit{e}.\textit{g}.}
\newcommand{\aka}{\textit{a}.\textit{k}.\textit{a}.}

\newcommand{\pa}{patch aggregation}

\title{\our{}: Masked Image Modeling with \\ Vector-Quantized Visual Tokenizers}

\author{%
{Zhiliang Peng$^{1}$\thanks{~Contribution during internship at Microsoft Research.},~~Li Dong$^{2}$,~~Hangbo Bao$^{2}$,~~Qixiang Ye$^{1}$,~~Furu Wei$^{2}$} \\
University of Chinese Academy of Sciences$^{1}$ \\
Microsoft Research$^{2}$ \\
\url{https://github.com/microsoft/unilm}
}

\begin{document}

\maketitle

\begin{abstract}
Masked image modeling (MIM) has demonstrated impressive results in self-supervised representation learning by recovering corrupted image patches. However, most existing studies operate on low-level image pixels, which hinders the exploitation of high-level semantics for representation models. In this work, we propose to use a semantic-rich visual tokenizer as the reconstruction target for masked prediction, providing a systematic way to promote MIM from pixel-level to semantic-level. Specifically, we propose vector-quantized knowledge distillation to train the tokenizer, which discretizes a continuous semantic space to compact codes. We then pretrain vision Transformers by predicting the original visual tokens for the masked image patches. 
Furthermore, we introduce a patch aggregation strategy which associates discrete image patches to enhance global semantic representation.
Experiments on image classification and semantic segmentation show that \our{} outperforms all compared MIM methods. On ImageNet-1K (224 size), the base-size \our{} achieves $85.5\%$ top-1 accuracy for fine-tuning and $80.1\%$ top-1 accuracy for linear probing. The large-size \our{} obtains $87.3\%$ top-1 accuracy for ImageNet-1K (224 size) fine-tuning, and $56.7\%$ mIoU on ADE20K for semantic segmentation. 
The code and pretrained models are available at \url{https://aka.ms/beitv2}.
\end{abstract}


\section{Introduction}
\label{sec:intro}
Masked image modeling (MIM), which greatly relieves the annotation-hungry issue of vision Transformers, has demonstrated great potential in learning visual representations~\citep{beit,mae}.
Given an image, the pretraining objective of MIM is to recover the masked patches so that rich context information is captured by the representation model. 
%
%
Taking BEiT~\citep{beit} as an example, each image has two views during pretraining, \ie{}, image patches, and visual tokens. The original image is first tokenized to discrete tokens. Randomly sampled image patches are then masked before being fed to vision Transformers. The pretraining objective is to recover the original visual tokens based on the corrupted image patches. The pretrained vision encoder can be deployed and finetuned on various downstream tasks by appending lightweight task layers.

Existing MIM approaches can be coarsely categorized to three according to the reconstruction targets: low-level image elements ($e.g.$, raw pixels; \citealt{mae,cim,swinv2}), hand-crafted features ($e.g.$, HOG features; \citealt{maskfeat}), and visual tokens; \citealt{beit,beit3,peco,splitmask,cae}.
However, all the reconstruction targets are about, explicitly or implicitly,  low-level image elements while underestimating high-level semantics.
%
In comparison, the masked words in language modeling~\citep{bert,unilm} are all about high-level semantics, which motivates us to tap the potential of MIM by exploiting semantic-aware supervision during pretraining.


In this work, we propose a self-supervised representation learning approach, termed \our{}, with the aim to improve MIM pretraining by constructing a semantic-aware visual tokenizer. 
Our approach is developed on the \beit{} method which is simple yet effective. 
The novelty lies in introducing the Vector-Quantized Knowledge Distillation (\vqkd{}) algorithm to discretize a semantic space.
The \vqkd{} encoder first converts the input image to discrete tokens according to a learnable codebook. The decoder then learns to reconstruct the semantic features encoded by a teacher model, conditioning on the discrete tokens. After training \vqkd{}, its encoder is used as a semantic visual tokenizer for \beit{} pretraining, where the discrete codes serve as supervision signals.

Considering the discreteness of tokens, we further introduce a patch aggregation strategy which explicitly encourages the \cls{} token to associate all patches~\citep{condenser}. Such a strategy resolves the issue that MIM put patch reconstruction the first place which diminishes learning global image representations.
As a result, \our{} improves the capacity of learned image representation, as supported by the linear probing experiments.
Moreover, the enhanced representations also boosts the performance of other tasks.

\begin{figure}[t]
\centering
\subfloat{
\includegraphics[width=0.42\textwidth]{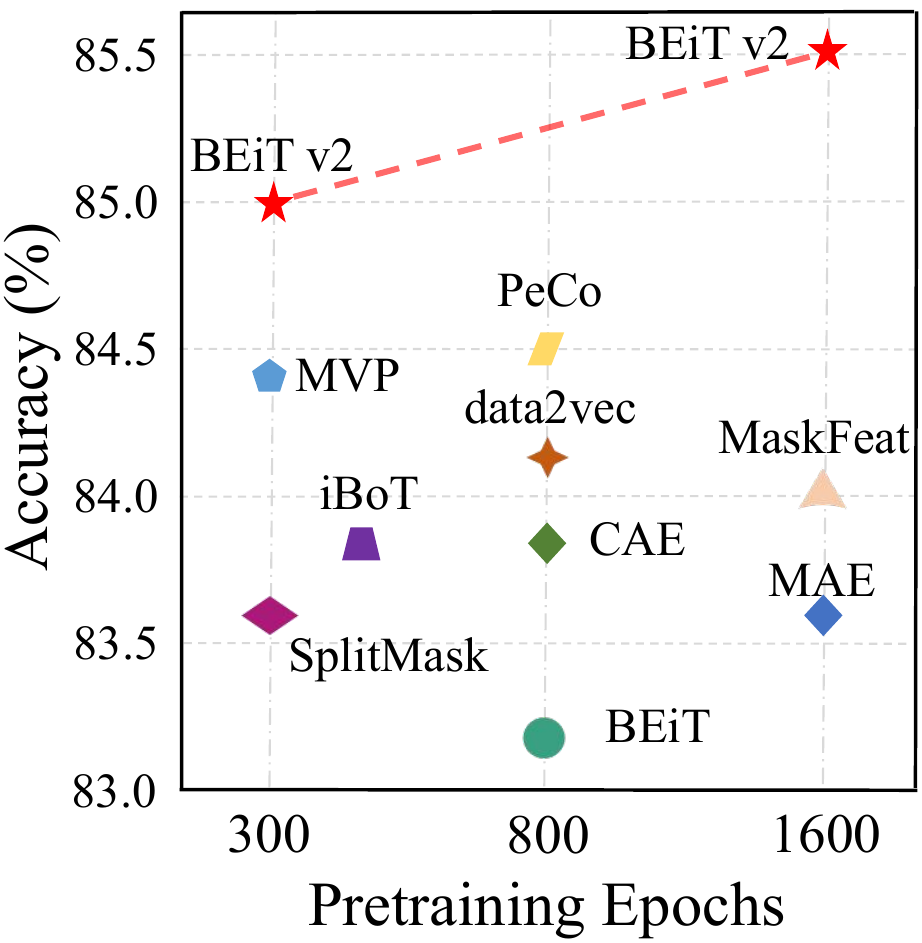}
}
\hspace{0.2in}
\subfloat{
\includegraphics[width=0.42\textwidth]{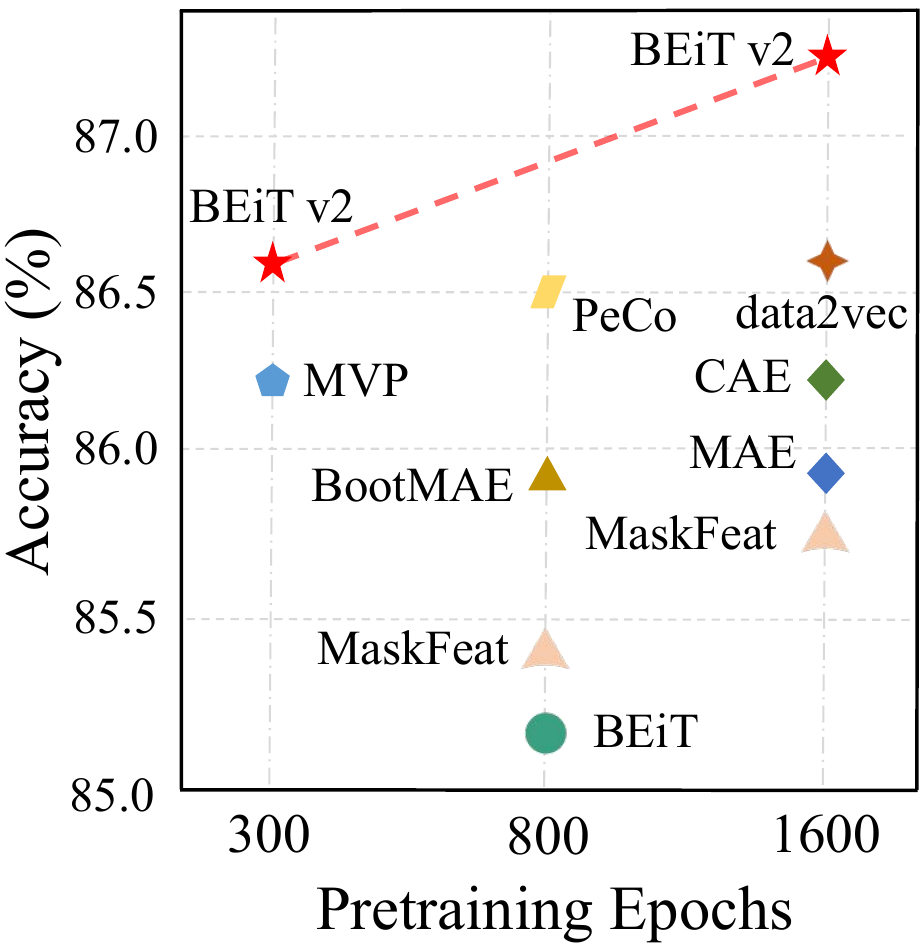}
}
\caption{ Top-1 fine-tuning accuracy on ImageNet (224 size). \textbf{Left}: ViT-B/16. \textbf{right}: ViT-L/16.}
\label{fig:intro:in1k}
\end{figure}

We conduct self-supervised learning on ImageNet-1k for both base- and large-size vision Transformers, which are evaluated on downstream tasks, \eg{}, image classification, linear probing, and semantic segmentation. As shown in Figure~\ref{fig:intro:in1k}, \our{} outperforms previous self-supervised learning algorithms by a large margin on ImageNet fine-tuning, \eg{}, improving over \beit{}~\citep{beit} by about two points for both ViT-B/16 and ViT-L/16. 
\our{} outperforms all compared MIM methods on ImageNet linear probing while achieving large performance gains on ADE20k for semantic segmentation.

The contributions of this work are summarized as follows:
\begin{itemize}[leftmargin=1.5em]
\item We propose vector-quantized knowledge distillation, promoting masked image modeling from pixel-level to semantic-level for self-supervised representation learning.

\item We introduce a patch aggregation strategy, which enforces global structure given discrete semantic tokens, and improves the performance of learned representations.

\item We conduct extensive experiments on downstream tasks including ImageNet fine-tuning, linear probing, and semantic segmentation. Experimental results show that the proposed approach significantly improves performance across model sizes, training steps, and downstream tasks.
\end{itemize}

\section{Methodology}
\label{sec:methods}

\our{} inherits the masked image modeling framework defined by \beit{}~\citep{beit}, which uses a visual tokenizer to convert each image to a set of discrete visual tokens. The training target is to recover the masked visual tokens, each of which corresponds to an image patch. In Section~\ref{sec:vqkd}, we introduce a vector-quantized knowledge distillation algorithm, which is used to train a visual tokenizer. In Section~\ref{sec:beitv2}, we employ the visual tokenizer for \beit{} pretraining under the help of the patch aggregation strategy. 

\subsection{Image Representation}
\label{sec:input:rep}

The vision Transformers (ViTs;~\citealt{vit}) are employed as the backbone networks to obtain image representations. The input image $\vx \in \R^{H \times W \times C}$ is reshaped to $N={HW}/{P^2}$ patches $\{\vx^{p}_{i} \}_{i=1}^{N}$, where $\vx^{p} \in \R^{N \times (P^2 C)}$ and $(P, P)$ is the patch size. In experiments, each $224 \times 224$ image is split to a $14 \times 14$ grid of image patches, where each patch is $16 \times 16$. The image patches $\{\vx^{p}_{i} \}_{i=1}^{N}$ are then flattened and linearly projected to input embeddings for Transformers. The encoding vectors are denoted as $\{\vh_{i} \}_{i=1}^{N}$, which corresponds to $N$ image patches.

\subsection{Training Visual Tokenizer}
\label{sec:vqkd}

\begin{figure}[t]
\begin{center}
\begin{tabular}{c}
\includegraphics[width=1\textwidth]{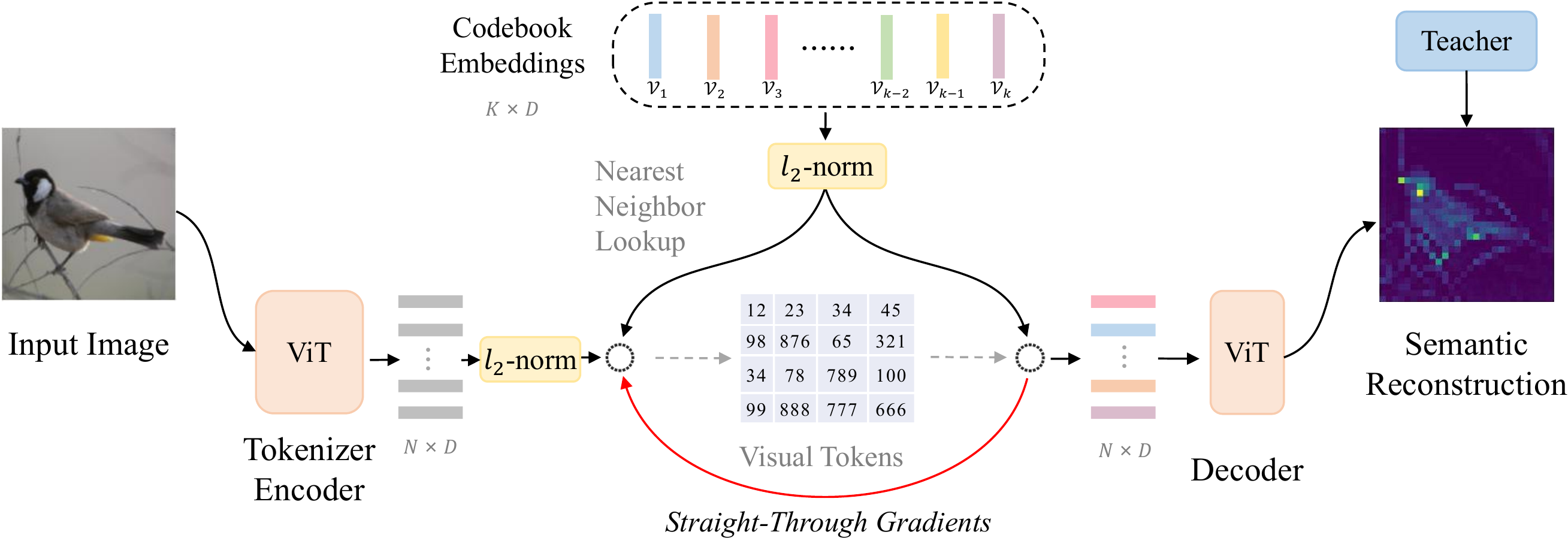}
\end{tabular}
\end{center}
\caption{Pipeline for visual tokenizer training. After training, each image is converted to discrete visual tokens.
}
\label{fig:vqkd}
\end{figure}
We propose vector-quantized knowledge distillation (\vqkd{}) to train the visual tokenizer, Figure~\ref{fig:vqkd}, where the visual tokenizer and the decoder are two vital modules.

The visual tokenizer maps an image to a sequence of visual tokens, \aka{}, discrete codes.
To be specific, an image $\vx$ is tokenized to $\vz = [ z_1, z_2, \cdots, z_{N} ] \in \gV^{ (H/P) \times (W/P) } $, where the visual vocabulary (\aka{}, codebook) $\gV \in \R^{K\times D}$ contains $K$ discrete codebook embeddings.

The tokenizer is consist of a vision Transformer encoder, and a quantizer.
The tokenizer first encodes the input image to vectors.
Then, the vector quantizer looks up the nearest neighbor in the codebook for each patch representation $\vh_{i}$.
Let $\{ \vv_1, \vv_2, \cdots, \vv_{K} \}$ denote the codebook embeddings.
For the $i$-th image patch, its quantized code is calculated as
\begin{align}
\vz_i = \argmin_{j} || \ell_2( \vh_i ) - \ell_2( \vv_j )||_2,
\label{eq:vq_distance}
\end{align}
where $j \in \{1, 2, \cdots, K\}$ and $\ell_2$ normalization is used for codebook lookup~\citep{vitvqgan}. 
The above distance is equivalent to finding codes according to cosine similarity.

After quantizing the image to visual tokens, we feed the $\ell_2$-normalized codebook embeddings $\{ \ell_2(\vv_{z_i}) \}_{i=1}^{N}$ to the decoder.
The decoder is also a multi-layer Transformer.
The output vectors $\{ \vo_i \}_{i=1}^{N}$ aim at reconstructing the semantic features of a teacher model, \eg{}, DINO~\citep{dino}, and CLIP~\citep{clip}.
Let $\vt_i$ denote the teacher model's feature vector of the $i$-th image patch.
During training, we maximize the cosine similarity between the decoder output $\vo_i$ and the teacher guidance $\vt_i$.

Because the quantization process (Equation~\ref{eq:vq_distance}) is non-differentiable,  the gradients are directly copied from the decoder input to the encoder output~\citep{vqvae},  Figure~\ref{fig:vqkd}, to back-propagate gradients to the encoder. Intuitively, the quantizer looks up the nearest code for each encoder output, while the gradients of codebook embeddings indicate useful optimization directions for the encoder.

The training objective of \vqkd{} is defined as
\begin{align}
\mathrm{max} \sum_{x \in \gD} \sum_{i=1}^N
\cos{( \vo_i , \vt_i )}
- ||\mathrm{sg}[\ell_2( \vh_i )] - \ell_2( \vv_{z_i} )||_2^2
- ||\ell_2( \vh_i ) - \mathrm{sg}[\ell_2( \vv_{z_i} )]||_2^2,
\label{eq:vqkd_objective}
\end{align}
where $\mathrm{sg}[\cdot]$ stands for the stop-gradient operator which is an identity at the forward pass while having zero gradients during the backward pass. $\gD$ represents the image data used for tokenizer training.

\paragraph{Improving codebook utilization.}
A common issue of vector quantization training is codebook collapse. In other words, only a small proportion of codes are used. Empirical strategies ~\citep{vqvae,vitvqgan} can be used to alleviate this issue. Equation~\ref{eq:vq_distance} shows that we compute the $\ell_2$-normalized distance to find the nearest code while reducing the dimension of codebook embedding space to $32$-d. The low-dimensional codebook embeddings are mapped back to higher-dimensional space before being fed to the decoder. 
Exponential moving average~\citep{vqvae} is employed to update the codebook embeddings. Exponential moving average tends to be more stable for \vqkd{} training.

\subsection{Pretraining \our{}}
\label{sec:beitv2}

\begin{figure}[t]
\begin{center}
\begin{tabular}{c}
\includegraphics[width=1\textwidth]{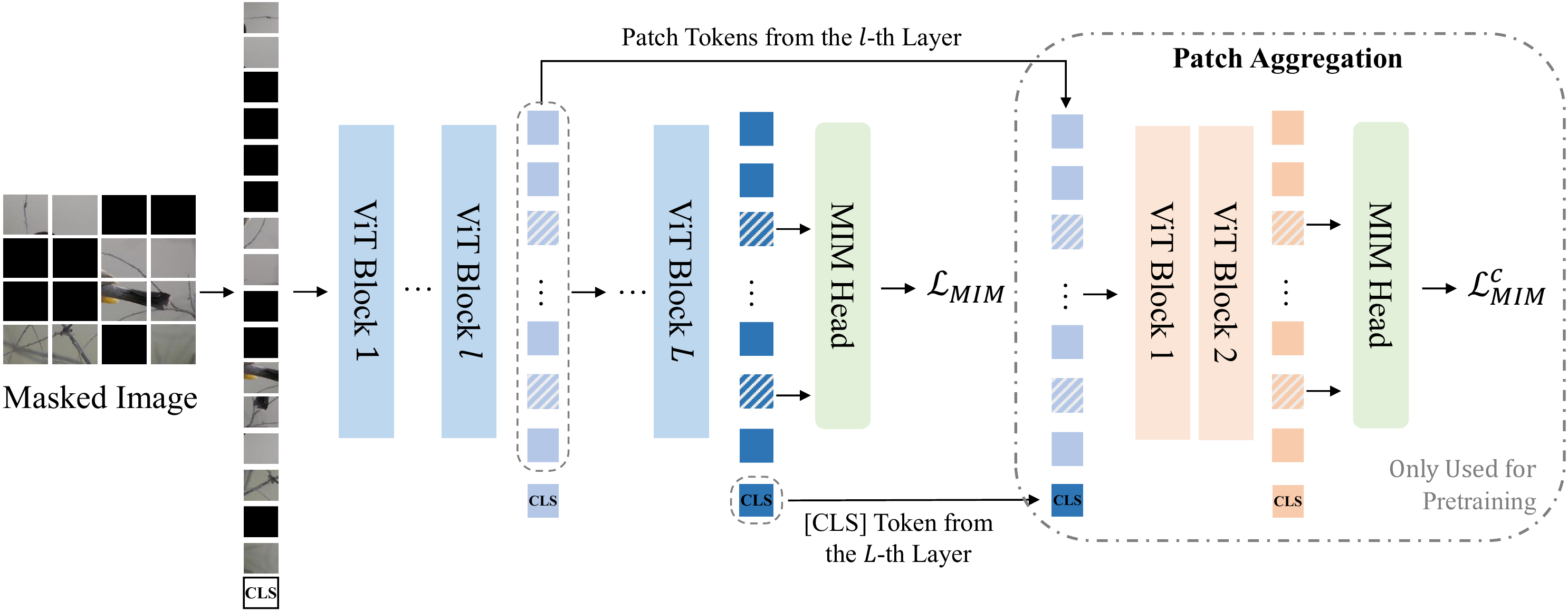}
\end{tabular}
\end{center}
\caption{The MIM framework equipped with \pa{}. The pretraining loss is the summation of $\mathcal{L}_{\text{MIM}}$ and $\mathcal{L}_{\text{MIM}}^{c}$. The loss term $\mathcal{L}_{\text{MIM}}^{c}$ explicitly encourages the \cls{} token to aggregate patch information to global representations.
}
\label{fig:beitv2}
\end{figure}

We follow the MIM setup in \beit{}~\citep{beit} to pretrain vision Transformers for image representations. Given an input image $x$, around 40\% image patches are block-wisely chosen and masked. The masked position is termed as $\gM$. 
%
Then, a shared learnable embedding $\ve_{[\text{M}]}$ is used to replace the original image patch embeddings $\ve^p_{i}$ if $i \in \gM$:
$\vx_{i}^{\gM} = \delta(i \in \gM) \odot \ve_{[\text{M}]} + (1 - \delta(i \in \gM)) \odot \vx^{p}_{i}$, where $\delta(\cdot)$ is the indicator function. Subsequently, we prepend a learnable \cls{} token to the input, \ie{}, $[\ve_{\texttt{CLS}}, \{ \vx_{i}^{\gM} \}_{i=1}^{N}]$, and feed them to the vision Transformer. The final encoding vectors are denoted as $\{\vh_{i} \}_{i=0}^{N}$, where $\vh_{0}$ is for the \cls{} token.

Next, we instantiate the MIM head as a simple fully-connection layer, and then use it to predict the visual tokens of the masked positions based on the corrupted image $\vx^{\gM}$. 
For each masked position $\{ \vh_{i} : i \in \gM \}_{i=1}^{N}$, a softmax classifier predicts the visual tokens $p( \vz_i | \vh_i ) = \softmax_{\vz_i} (\mW_{c} \vh_{i} + \vb_{c})$, where $\mW_{c}, \vb_{c}$ respectively mean weights and biases of the MIM head. The visual tokens are obtained by the tokenizer trained in Section~\ref{sec:vqkd}, which provides supervisions for the MIM self-supervised learning procedure.
The training loss of MIM is defined as
\begin{align}
\mathcal{L}_{\rm{MIM}} = - \sum_{\vx \in \gD} \sum_{i \in \gM} \mathrm{log} \ p( \vz_i | \vx^{\gM}_i ),
\label{eq:mim_objective}
\end{align}
where $\vz_i$ denotes the visual tokens of the original image, and $\gD$ the pretraining images. Notice that the number of visual tokens is the same as the number of image patches in this work.

\paragraph{Pretraining global representation.}
Inspired by~\citep{condenser}, we pretrain the \cls{} token for global image representation. The goal is to mitigate the discrepancy between patch-level pretraining and image-level representation aggregation. As illustrated in Figure~\ref{fig:beitv2}, a representation bottleneck is constructed to encourage the \cls{} token to gather information as much as possible.
For a $L$-layer Transformer, let $\{\vh_{i}^{l}\}_{i=1}^{N}$ denote the $l$-th layer's output vectors, where $l\in\{1, 2, \cdots,L\}$. To pretrain the last layer's \cls{} token $\vh_{\texttt{CLS}}^{L}$, we concatenate it with the intermediate $l$-th layer's patch vectors $\{\vh_{i}^{l}\}_{i=1}^{N}$, \ie{}, $\mS = [\vh_{\texttt{CLS}}^{L}, \vh_{1}^{l}, \cdots, \vh_{N}^{l}]$.
We then feed $\mS$ to a shallow (\eg{}, two layers) Transformer decoder and conduct masked prediction again,
\ie{}, $p( \vz | \mS ) = \softmax_{\vz} (\mW_{c} \mS + \vb_{c})$.
Notice that the parameters are shared for both MIM heads and the MIM loss is also computed at mask positions as in Equation~\ref{eq:mim_objective}. Accordingly, the final training loss is defined as the summation of two terms, \ie{}, the original loss at the $L$-th layer, and the shallow Transformer decoder's MIM loss. 
Overall framework refers to  Appendix~\ref{app:overall_framework}.

Intuitively, the model favors pushing the global information to $\vh_{\texttt{CLS}}^{L}$, because the model tends to fully utilize the parameters from ($l+1$)-th layer to $L$-th layer, to decrease the additional MIM loss.
The information-flow bottleneck encourages the \cls{} token towards more reliable global representations than its untrained counterparts.
Moreover, the enhanced representations also facilitate various downstream tasks.
Notice that the newly added shallow decoder is only used to pretrain the \cls{} token, which is discarded after pretraining.

\section{Experiments}
\label{sec:exp}

The pretrained models are evaluated on image classification and semantic segmentation tasks. For image classification, the models are trained on ImageNet-1K ~\citep{imagenet} and evaluated by (1) top-1 accuracy about fine-tuning and (2) top-1 accuracy about linear probing (only fine-tuning the classification head). For semantic segmentation, experiments are conducted on the ADE20K dataset~\citep{ade20k} and the performance is evaluated using the mIoU protocol.

\begin{table}[!t]
\centering
\caption{
Fine-tuning results of image classification and semantic segmentation on ImageNet-1K and ADE20k. UperNet~\citep{upernet} is used as the task layer for semantic segmentation with single-scale (512 size) input.
}
\label{tbl:results}
\begin{tabular}{@{}lccc@{}}
\toprule
\bf \multirow{2}{*}{Methods} & \bf Pretraining  & \bf  ImageNet & \bf ADE20k \\
\bf  & \bf Epochs & \bf Top-1 Accuracy(\%)  & \bf mIoU(\%) \\
\midrule
\multicolumn{4}{l}{\textit{Base-size models (ViT-B/16)}} \\
\beit{}~\citep{beit}   & 300  & 82.9 & 44.7 \\
CAE~\citep{cae}         & 300  & 83.6 & 48.3 \\
SplitMask~\citep{splitmask} & 300  & 83.6 & 45.7 \\
MaskFeat~\citep{maskfeat} & 300  & 83.6 & N/A \\
PeCo~\citep{peco} & 300   & 84.1 & 46.7 \\
MVP~\citep{mvp}  & 300   & 84.4 & 52.4 \\
\bf \our{} (ours)  & 300  & \bf 85.0 & \bf 52.7 \\ 
\multicolumn{4}{l}{\textit{Base-size models (ViT-B/16) + pretrain longer}} \\
\beit{}~\citep{beit}   & 800  & 83.2 & 45.6 \\
PeCo~\citep{peco}       & 800   & 84.5 & 48.5 \\
data2vec~\citep{data2vec} & 800   & 84.2 & N/A \\
MAE~\citep{mae}         & 1600   & 83.6 & 48.1 \\
CAE~\citep{cae}         & 1600   & 83.9 & 50.2 \\
\bf \our{} (ours)  & 1600   & \bf 85.5 & \bf 53.1 \\
~~$+$ Intermediate fine-tuning with ImageNet-21k  &    & \bf 86.5 & \bf 53.5 \\
\midrule
\multicolumn{4}{l}{\textit{Large-size models (ViT-L/16)}} \\
MaskFeat~\citep{maskfeat} & 300  & 84.4 & N/A \\
MVP~\citep{mvp}  & 300   & 86.3 & 54.3 \\
\bf \our{} (ours)  & 300  & \bf 86.6 & \bf 55.0 \\
\multicolumn{4}{l}{\textit{Large-size models (ViT-L/16) + pretrain longer}} \\
\beit{}~\citep{beit}  & 800   & 85.2 & 53.3 \\
MaskFeat~\citep{maskfeat} & 1600 & 85.7 & N/A \\
MAE~\citep{mae}          & 1600   & 85.9 & 53.6 \\
CAE~\citep{cae}          & 1600   & 86.3 & 54.7 \\
data2vec~\citep{data2vec}  & 1600   & 86.6 & N/A \\
\bf \our{} (ours)  & 1600   & \bf 87.3 & \bf 56.7 \\
~~$+$ Intermediate fine-tuning with ImageNet-21k  &    & \bf 88.4 & \bf 57.5 \\
\bottomrule
\end{tabular}
\end{table}

\subsection{Pretraining Setup}
\label{sec:exp:setup}

\paragraph{Visual tokenizer training.}
We instantiate the visual tokenizer of \vqkd{} as ViT-B/16 for both base- and large-size \our{} pretraining.
The decoder network is a three-layer standard Transformer, which has the same dimension and number of attention heads as the tokenizer encoder.
The OpenAI {CLIP-B/16}~\citep{clip} is employed as the teacher model and train \vqkd{} on ImageNet-1k with 224$\times$224 resolution.
Notice that we use the same base-size teacher to train the visual tokenizer for both base- and large-size pretraining.
The code size $K$ is set as 8192 and code dimension $D$ as 32 by default.
%
Refer to Appendix~\ref{app:vqkd} for more training details.

\paragraph{Masked image modeling.}
We follow the settings used in BEiT~\citep{beit} pretraining and use ImageNet-1K without labels as the pretraining data for self-supervised learning.
The input image resolution is set as 224x224 during pretraining.
The pretrained base- and large-size vision Transformers~\citep{vit} with $16 \times 16$ patch size are denoted as ViT-B/16 and ViT-L/16, respectively.
For the \pa{} strategy, we set $l=9$ for ViT-B/16, $l=21$ for ViT-L/16, and the depth as 2 by default.
A block-wise masking mechanism is adopted under the mask ratio of 40\% (\ie{}, about 75 image patches).
More pretraining details can be found in Appendix~\ref{app:pretrain}.

\begin{table*}[t]
\centering
\small
\begin{minipage}{2.5in}
\centering
\caption{Top-1 accuracy of linear probing on ImageNet-1k. All methods are based on ViT-B/16 pretrained for 300 epochs except MAE for 1600 epochs. 
}
\label{tbl:results:linear}
\resizebox{\linewidth}{!}{ 
\begin{tabular}{@{}lc@{}}
\toprule
\bf Methods & \bf Linear Probe \\
\midrule
\beit{}~\citep{beit}  & 56.7 \\
CAE~\citep{cae} & 64.1 \\
MAE~\citep{mae} & 67.8 \\
MVP~\citep{mvp} & 75.4 \\
MoCo v3~\citep{mocov3} & 76.7 \\
\our{} (ours) & \bf 80.1 \\
\bottomrule
\end{tabular}
} 
\end{minipage}
\hfill
\begin{minipage}{2.6in}
\centering
\caption{Robustness evaluation on three ImageNet variants~\citep{adversarial2021,rendition2021,sketch2019}.
}
\label{tbl:results:robust}
\resizebox{\linewidth}{!}{ 
\begin{tabular}{@{}lccc@{}}
\toprule
\bf \multirow{2}{*}{Methods} & \bf ImageNet & \bf ImageNet & \bf ImageNet \\
 & \bf Adversarial & \bf Rendition & \bf Sketch \\
\midrule
\multicolumn{4}{l}{\textit{ViT-B/16}} \\
MAE  & 35.9 & 48.3 & 34.5 \\
\our{}  & \textbf{54.4} & \textbf{61.0} & \textbf{45.6} \\
\midrule
\multicolumn{4}{l}{\textit{ViT-L/16}} \\
MAE &  57.1 & 59.9 & 45.3 \\
\our{}  & \textbf{69.0} & \textbf{69.9} & \textbf{53.5} \\
\bottomrule
\end{tabular}
} 
\end{minipage}
\vspace{-2em}
\end{table*}

\subsection{Image Classification}
\label{sec:image:cls}

Both the fine-tuning accuracy and linear probing accuracy are evaluated on ImageNet-1k by default. The models are also evaluated on several ImageNet variants to demonstrate their favorable generalization ability.

\paragraph{Fine-tuning setup.}
We follow the protocol proposed in BEiT~\citep{beit} to fine-tune the pretrained \our{} model (see Appendix~\ref{app:finetune:cls} for more details).
%
In Table~\ref{tbl:results}, we report the top-1 fine-tuning accuracy results and compare \our{} with recent MIM methods.

From Table~\ref{tbl:results}, 
base-size \our{} with a 300-epoch pretraining schedule reaches 85.0\% top-1 accuracy, 
which outperforms \beit{}, CAE, SplitMask and PeCo by 2.1\%, 1.4\%, 1.4\% and 0.9\% respectively. 
Compared with masked distillation methods, like  MVP, \our{} also shows superiority.
Furthermore, with a longer pretraining schedule, \our{} achieves 85.5\% top-1 accuracy, developing a new state of the art on ImageNet-1K among self-supervised methods.
Meanwhile, \our{} using ViT-L/16 with 300 epochs reaches 86.6\% top-1 accuracy, which is comparable to data2vec with 1600 epochs. A longer pretraining schedule further boosts the performance to 87.3\%.

{\color{black}
Following \beit{}, we add an intermediate fine-tuning phase between the pretraining stage and the fine-tuning stage. Only the intermediate fine-tuning phase uses the ImageNet-21k dataset.
As shown in Table~\ref{tbl:results}, we find that intermediate fine-tuning achieves about 1\% performance gain on image classification for both base- and large-size models. Refer to Appendix~\ref{app:large_scale} for more results of intermediate fine-tuning. 
}

\paragraph{Linear probing.} Keeping the backbone model frozen and training a linear classification head atop the image-level representations, linear probing
has been a widely considered measure for self-supervised learning.
We average the patch tokens as the global representation for the models without \pa{}. Otherwise, we consider the \cls{} token as the global representation.
%
Table~\ref{tbl:results:linear} presents the top-1 accuracy for linear probing and compares \our{} with recent methods including \beit{}, CAE, MAE, MVP and MoCo v3.
All the compared methods are based on ViT-B/16 and pretrained for 300 epochs except MAE for 1600 epochs. \our{} respectively outperforms \beit{}, CAE and MVP by 23.4\%, 16.0\% and 4.7\%. \our{} also outperforms MoCo v3, which learns a global representation through a contrastive learning fashion. The comparisons indicate that the representation models learned by \our{} enjoy higher adaptation capability.

\paragraph{Robustness evaluation.}
We evaluate the robustness of \our{} on various ImageNet validation sets, $\ie{}$, ImageNet-Adversarial~\citep{adversarial2021}, ImageNet-Rendition~\citep{rendition2021} and ImageNet-Sketch~\citep{sketch2019}.
As shown in Table~\ref{tbl:results:robust}, compared with MAE~\citep{mae}, \our{} achieves dramatic gains across datasets, demonstrating the superiority of the proposed method in terms of model generalization.

\subsection{Semantic Segmentation}
\label{sec:results:seg}

Semantic segmentation is a dense prediction task, which generates class label for each pixel of the input image.
Following the setting proposed in \beit{}~\citep{beit}, we conduct experiments on ADE20K benchmark~\citep{ade20k}, which includes 25K mages and 150 semantic categories. 
We use UperNet~\citep{upernet} task layer and fine-tune the model for 160K iterations with the input resolution $512 \times 512$.
Refer to Appendix~\ref{app:finetune:seg} for details.
Table~\ref{tbl:results} shows that \our{} significantly outperforms previous self-supervised methods.
Moreover, using the ViT-L/16 model, the performance can reach 56.7, which builds a new state-of-the-art for masked image modeling on ADE20k.

\subsection{Analysis}
\label{sec:analysis}

\begin{table}[t]
\centering
\caption{Ablation studies under \vqkd{} settings. ``Base\&1x768x12'' denotes that the encoder network is ViT-Base while the decoder is a Transformer with depth 1, dimensions 768, and head 12. ``Reconst. Loss'' is the reconstruction loss of \vqkd{}. Reconstruction loss and codebook usage are measured on the validation set. After 300 epochs of pretraining, our method reports the top-1 fine-tuning accuracy and linear probing accuracy on ImageNet-1k, and mIoU on ADE20k.
The default setting is highlighted in \colorbox{baselinecolor}{gray}.
}
\label{tbl:ablation:vqkd}
\small
\begin{tabular}{c c c c c c c}
\toprule
 \bf \vqkd{} & \bf \multirow{2}{*}{Codebook} & \bf Reconst. & \bf Codebook & \bf ImageNet & \bf ImageNet & \bf \multirow{2}{*}{ADE20k} \\
 \bf Architecture &  & \bf Loss & \bf Usage & \bf Fine-tuning & \bf Linear Probe &  \\
\midrule
 Small \& 1x384x6 & \multirow{4}{*}{8192$\times$32}  & 0.183 & 100\% & 84.3 & 76.0 & 51.0 \\
 Base \& 1x768x12 &  & 0.164 & 100\% & 84.7 & 78.5 & 51.8 \\
 Base \& 3x768x12 &  & 0.145 & 95\%  & \baseline{84.7} & \baseline{77.9} & \baseline{51.9} \\
 Base \& 6x768x12 &  & 0.136 & 77\%  & 84.6 & 63.0 & 50.1 \\
 \midrule
\multirow{2}{*}{Base \& 3x768x12} & 8192$\times$16  & 0.145 & 100\% & 84.7 & 76.7 & 51.7 \\
 &  8192$\times$64 & 0.148 & 67\% & 84.7 & 77.6 & 51.6 \\
\bottomrule
\end{tabular}
\end{table}

\paragraph{Visual tokenizer training.}
We investigate the impact of \vqkd{} on \our{} in terms of the model architecture and codebook size and report the results in Table~\ref{tbl:ablation:vqkd}. 
ViT-B/16 without the \pa{} strategy is used as the baseline model, which is pretrained for 300 epochs.
As shown in Table~\ref{tbl:ablation:vqkd}, we find that a deeper decoder of \vqkd{} obtains better reconstruction, but lower codebook usage and downstream task performance.
Reducing dimension for codebook lookup improves codebook utilization~\citep{vitvqgan}.

\begin{table*}[t]
\centering
\caption{Ablation studies for \pa{} strategy. {\bf $l$-th Layer} denotes path tokens from the $l$-th layer of the backbone. {\bf Head Depth} means the \pa{} head depth. {\bf Shared MIM Head} means whether we share the MIM head parameters or not. Default settings are in \colorbox{baselinecolor}{gray}.
}
\label{tbl:ablation:cls_pt}
\begin{tabular}{@{}c c c c c c@{}}
\toprule
\bf $l$-th & \bf Head & \bf Shared & \bf ImageNet & \bf ImageNet & \bf \multirow{2}{*}{ADE20k}  \\
\bf Layer & \bf Depth & \bf MIM Head & \bf Fine-tuning  & \bf Linear Probe &  \\
\midrule
\multicolumn{6}{l}{\textit{{~~~Without \pa{}}}} \\
- & - & - & {84.7} & {77.9} & {51.9} \\
\midrule
\multicolumn{6}{l}{\textit{{~~~With \pa }}} \\
9 & 2 & \cmark &\bf \baseline{85.0} & \bf \baseline{80.1} & \baseline{52.7} \\
9 & 2 & \xmark & 84.8 & 79.5 & 51.9 \\
9 & 1 & \cmark & 84.8 & 78.9 & 51.7 \\
9 & 3 & \cmark & 84.7 & 78.1 & 52.0 \\
6 & 2 & \cmark & 84.9 & 77.5 & \bf 53.1 \\
11 & 2 & \cmark & 84.5 & 69.4 & 51.8 \\
\bottomrule
\end{tabular}
\end{table*}

\paragraph{Patch aggregation strategy.}  
Table~\ref{tbl:ablation:cls_pt} presents the ablation studies of the \pa{} strategy. 
The shallower head (i.e., 1/2-layer) performs better than the deeper head (i.e., 3-layer), suggesting the shallower head pays more attention to the input \cls{} token than the deeper head.
Moreover, the proposed method outperforms the baseline variant without \pa{} strategy. The improvement of linear probe indicates better image-level representations.
In addition, the results indicate that sharing the MIM head improves downstream performance.

\begin{table*}[t]
\centering
\caption{Comparisons between different \vqkd{} targets. We also report the fine-tuning results of \vqkd{} target models.
}
\label{tbl:results:vqkd_teacher}
\begin{tabular}{@{}lcc@{}}
\toprule
\bf {\vqkd{} Targets} & \bf {ImageNet} & \bf {ADE20k} \\
\midrule
\multicolumn{3}{l}{\textit{Pretrain 300 epochs}} \\
DINO &  84.4 & 49.2 \\
CLIP  &  {85.0}  &  {52.7} \\
\multicolumn{3}{l}{\textit{Pretrain 1600 epochs}} \\
CLIP  & \bf {85.5} & \bf {53.1} \\
\midrule
\multicolumn{3}{l}{\textit{\gc{Performance of VQ-KD target models}}} \\
\gc{DINO}  & \gc{83.6} & \gc{46.8} \\
\gc{CLIP}  & \gc{84.9}  & \gc{-} \\
\multicolumn{3}{l}{\textit{\gc{Performance of VQ-KD encoder model}}} \\
\gc{\vqkd{} encoder (CLIP as target)}  & \gc{83.6}  & \gc{-} \\
\bottomrule
\end{tabular}
\end{table*}

\paragraph{\vqkd{} targets.} 
In Table~\ref{tbl:results:vqkd_teacher}, we report the results about \vqkd{}s are trained under the supervision of DINO~\citep{dino} and CLIP~\citep{clip}. DINO is pretrained solely on ImageNet-1k while CLIP is pretrained on 400M image-text pairs datasets in house.
%
We also directly fine-tune the official base-size checkpoints and report the results in Table~\ref{tbl:results:vqkd_teacher}.
One can see that when using DINO as the teacher model, \our{} respectively reaches 84.4\% and 49.2\% on ImageNet and ADE20k, outperforming DINO itself by a large margin.
When using CLIP as the teacher model, \our{} can get consistent improvements, demonstrating the scalability of the proposed \vqkd{}.
In addition, we directly fine-tune the \vqkd{} encoder on ImageNet. The results show that transfer performance of the \vqkd{} encoder is lower than the teacher model.
After performing masked image modeling, the pretrained model outperforms both the teacher model and the visual tokenizer encoder.
It demonstrates the superiority of the proposed method for self-supervised learning.

\begin{figure}[t]
\begin{center}
\begin{tabular}{c}
\includegraphics[width=0.99\textwidth]{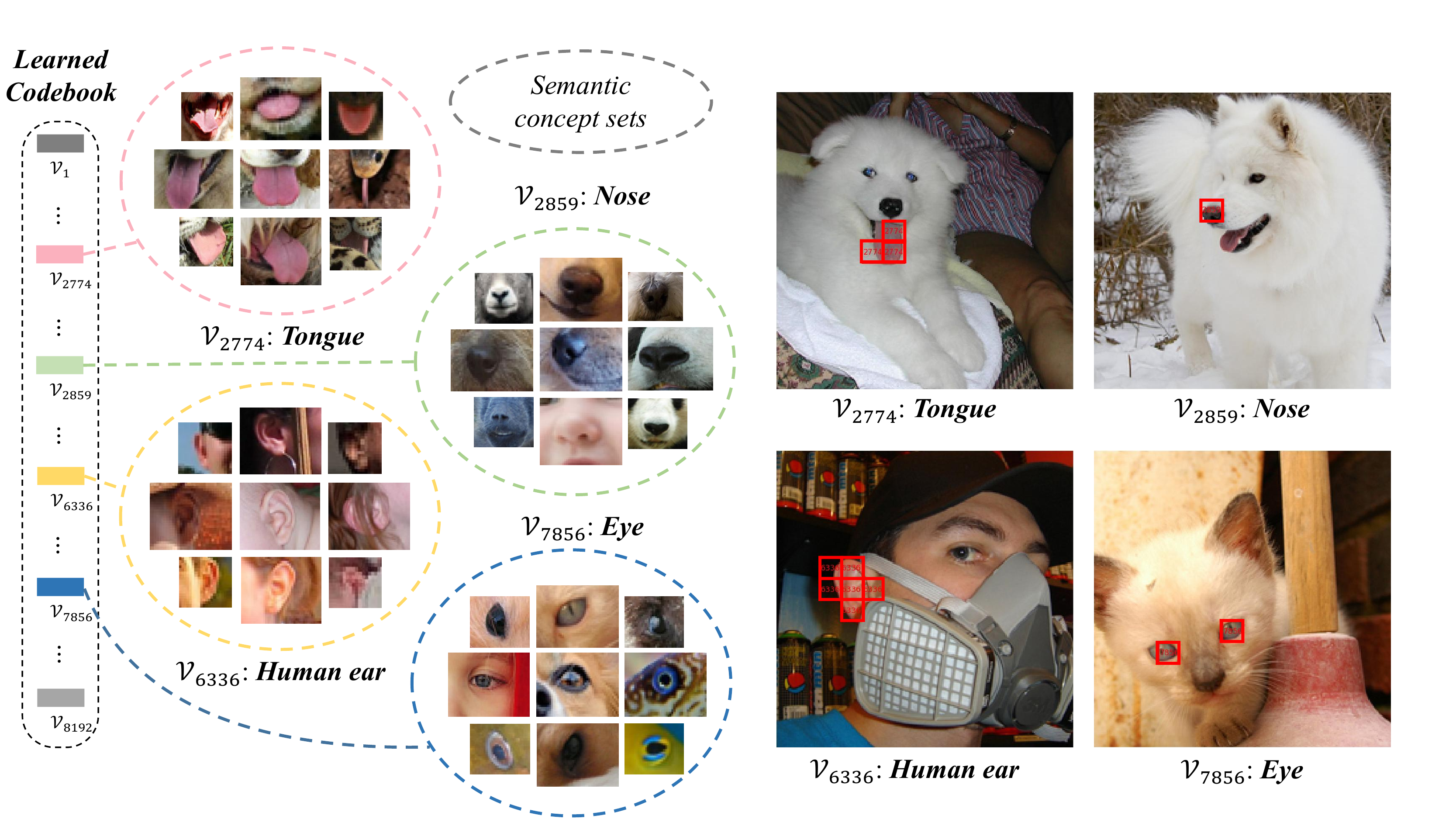}
\end{tabular}
\end{center}
\caption{
Visualization of semantic concepts corresponding to the learned codebook.
}
\label{fig:vis_codebook}
\end{figure}

\paragraph{Visualization of codebook.}
We utilize the proposed \vqkd{} to calculate discrete codes about the ImageNet-1k validation set. Image patches are grouped according to their corresponding codes.
Figure~\ref{fig:vis_codebook} shows that the grouped image patches represent explicit semantics.
For instance, the image patches corresponding to code $7856$ are about ``eyes'' of human, cat, dog, fish and snake.
Refer to Appendix~\ref{app:vis_code}) for more examples.
The introduction of codebook and feature quantization reduces the sensitiveness to the change of image details while facilitates exploitation of high-level semantics for representation models. 
\vqkd{} compresses and quantizes the continuous feature values to a codebook, which constructs a discrete semantic space. The dimensionality of such a semantic space is  significantly lower than that of the original continuous feature space. This reduces difficulty of masked patch reconstruction and alleviates the curse of dimensionality in the pretraining phase.

%


\section{Related Work}
\label{sec:related:work}

\paragraph{Visual tokenizer.}
VQ-VAE~\citep{vqvae} converts an image into a sequence of discrete codes and then reconstructs the input image based on discrete codes.
DALL-E~\citep{dalle} uses the Gumbel-softmax relaxation for quantization instead of the nearest neighbor lookup in \vae{}.
VQGAN~\citep{vqgan} and ViT-VQGAN~\citep{vitvqgan} introduce Transformer block to train a better autoencoder to maintain fine details with adversarial and perceptual loss.
Moreover, ViT-VQGAN proposes factorized and $\ell_2$-normalized code for codebook learning.
In comparison, the proposed \vqkd{} aims at reconstructing semantic knowledge from the teacher rather than original pixels. So we can construct a highly compact semantic codebook for MIM.

\paragraph{Masked image modeling.}

The MIM method has achieved great success in language task~\citep{bert,unilm,unilm2}. Motivated by it, \beit{}~\citep{beit} mitigated the MIM method to computer vision tasks by recovering discrete visual tokens~\citep{dalle}. The prediction targets for MIM have been explored by many recent works. MAE~\citep{mae} treated MIM as a denoising pixel-level reconstruction task. Knowledge distillation~\citep{maskfeat,mvp} and self-distillation~\citep{data2vec} proposed to mimic the features provided by the teacher at the masked positions. PeCo~\citep{peco} regarded MoCo v3~\citep{mocov3} as the perceptual model in VQGAN training~\citep{vqgan}, to pursue a better tokenizer for \beit{} pretraining. 
Despite of the progress, most existing studies remain operating on low-level image pixels,
this work explores how to promote masked image modeling from pixel-level to semantic-level.

\section{Conclusion}

We proposed vector-quantized knowledge distillation (\vqkd{}) to train a visual tokenizer for vision Transformer pretraining.
\vqkd{} discretized a continuous semantic space that provides supervision for masked image modeling rather than relying on image pixels.
The semantic visual tokenizer greatly improved the \beit{} pretraining and significantly boosted the transfer performance upon downstream tasks, such as image classification, and semantic segmentation.
Moreover, a patch aggregation mechanism was introduced to explicitly encourage the model to produce global image representations, narrowing the gap between the patch-level pretraining and image-level representation aggregation.
In the future, we would like to learn a universal tokenizer that projects words and images into the same vocabulary, so that we can conduct masked prediction for vision-language pretraining.



\bibliographystyle{iclr2023_conference}
\bibliography{beit}

\begin{thebibliography}{30}
\providecommand{\natexlab}[1]{#1}
\providecommand{\url}[1]{\texttt{#1}}
\expandafter\ifx\csname urlstyle\endcsname\relax
  \providecommand{\doi}[1]{doi: #1}\else
  \providecommand{\doi}{doi: \begingroup \urlstyle{rm}\Url}\fi

\bibitem[Baevski et~al.(2022)Baevski, Hsu, Xu, Babu, Gu, and Auli]{data2vec}
Alexei Baevski, Wei-Ning Hsu, Qiantong Xu, Arun Babu, Jiatao Gu, and Michael
  Auli.
\newblock Data2vec: A general framework for self-supervised learning in speech,
  vision and language.
\newblock \emph{arXiv preprint arXiv:2202.03555}, 2022.

\bibitem[Bao et~al.(2020)Bao, Dong, Wei, Wang, Yang, Liu, Wang, Gao, Piao,
  Zhou, and Hon]{unilm2}
Hangbo Bao, Li~Dong, Furu Wei, Wenhui Wang, Nan Yang, Xiaodong Liu, Yu~Wang,
  Jianfeng Gao, Songhao Piao, Ming Zhou, and Hsiao{-}Wuen Hon.
\newblock {UniLMv2}: Pseudo-masked language models for unified language model
  pre-training.
\newblock In \emph{Proceedings of the 37th International Conference on Machine
  Learning, {ICML} 2020}, volume 119 of \emph{Proceedings of Machine Learning
  Research}, pp.\  642--652. {PMLR}, 2020.

\bibitem[Bao et~al.(2022)Bao, Dong, Piao, and Wei]{beit}
Hangbo Bao, Li~Dong, Songhao Piao, and Furu Wei.
\newblock {BE}i{T}: {BERT} pre-training of image {Transformers}.
\newblock In \emph{International Conference on Learning Representations}, 2022.

\bibitem[Caron et~al.(2021)Caron, Touvron, Misra, J\'egou, Mairal, Bojanowski,
  and Joulin]{dino}
Mathilde Caron, Hugo Touvron, Ishan Misra, Herv\'e J\'egou, Julien Mairal,
  Piotr Bojanowski, and Armand Joulin.
\newblock Emerging properties in self-supervised vision {Transformers}.
\newblock \emph{arXiv preprint arXiv:2104.14294}, 2021.

\bibitem[Chen et~al.(2022)Chen, Ding, Wang, Xin, Mo, Wang, Han, Luo, Zeng, and
  Wang]{cae}
Xiaokang Chen, Mingyu Ding, Xiaodi Wang, Ying Xin, Shentong Mo, Yunhao Wang,
  Shumin Han, Ping Luo, Gang Zeng, and Jingdong Wang.
\newblock Context autoencoder for self-supervised representation learning.
\newblock \emph{arXiv preprint arXiv:2202.03026}, 2022.

\bibitem[Chen et~al.(2021)Chen, Xie, and He]{mocov3}
Xinlei Chen, Saining Xie, and Kaiming He.
\newblock An empirical study of training self-supervised vision {Transformers}.
\newblock \emph{ArXiv}, abs/2104.02057, 2021.

\bibitem[Devlin et~al.(2019)Devlin, Chang, Lee, and Toutanova]{bert}
Jacob Devlin, Ming{-}Wei Chang, Kenton Lee, and Kristina Toutanova.
\newblock {BERT:} pre-training of deep bidirectional {Transformers} for
  language understanding.
\newblock In \emph{Proceedings of the 2019 Conference of the North American
  Chapter of the Association for Computational Linguistics: Human Language
  Technologies}, pp.\  4171--4186. Association for Computational Linguistics,
  2019.

\bibitem[Dong et~al.(2019)Dong, Yang, Wang, Wei, Liu, Wang, Gao, Zhou, and
  Hon]{unilm}
Li~Dong, Nan Yang, Wenhui Wang, Furu Wei, Xiaodong Liu, Yu~Wang, Jianfeng Gao,
  Ming Zhou, and Hsiao{-}Wuen Hon.
\newblock Unified language model pre-training for natural language
  understanding and generation.
\newblock In \emph{Advances in Neural Information Processing Systems 32: Annual
  Conference on Neural Information Processing Systems 2019, NeurIPS 2019,
  December 8-14, 2019, Vancouver, BC, Canada}, pp.\  13042--13054, 2019.

\bibitem[Dong et~al.(2021)Dong, Bao, Zhang, Chen, Zhang, Yuan, Chen, Wen, and
  Yu]{peco}
Xiaoyi Dong, Jianmin Bao, Ting Zhang, Dongdong Chen, Weiming Zhang, Lu~Yuan,
  Dong Chen, Fang Wen, and Nenghai Yu.
\newblock {PeCo}: Perceptual codebook for bert pre-training of vision
  {Transformers}.
\newblock \emph{arXiv preprint arXiv:2111.12710}, 2021.

\bibitem[Dosovitskiy et~al.(2020)Dosovitskiy, Beyer, Kolesnikov, Weissenborn,
  Zhai, Unterthiner, Dehghani, Minderer, Heigold, Gelly, et~al.]{vit}
Alexey Dosovitskiy, Lucas Beyer, Alexander Kolesnikov, Dirk Weissenborn,
  Xiaohua Zhai, Thomas Unterthiner, Mostafa Dehghani, Matthias Minderer, Georg
  Heigold, Sylvain Gelly, et~al.
\newblock An image is worth 16x16 words: {Transformers} for image recognition
  at scale.
\newblock \emph{preprint arXiv:2010.11929}, 2020.

\bibitem[El-Nouby et~al.(2021)El-Nouby, Izacard, Touvron, Laptev, Jegou, and
  Grave]{splitmask}
Alaaeldin El-Nouby, Gautier Izacard, Hugo Touvron, Ivan Laptev, Herv{\'e}
  Jegou, and Edouard Grave.
\newblock Are large-scale datasets necessary for self-supervised pre-training?
\newblock \emph{arXiv preprint arXiv:2112.10740}, 2021.

\bibitem[Esser et~al.(2021)Esser, Rombach, and Ommer]{vqgan}
Patrick Esser, Robin Rombach, and Bjorn Ommer.
\newblock Taming {Transformers} for high-resolution image synthesis.
\newblock In \emph{CVPR}, pp.\  12873--12883, 2021.

\bibitem[Fang et~al.(2022)Fang, Dong, Bao, Wang, and Wei]{cim}
Yuxin Fang, Li~Dong, Hangbo Bao, Xinggang Wang, and Furu Wei.
\newblock Corrupted image modeling for self-supervised visual pre-training.
\newblock \emph{ArXiv}, abs/2202.03382, 2022.

\bibitem[Gao \& Callan(2021)Gao and Callan]{condenser}
Luyu Gao and Jamie Callan.
\newblock Condenser: a pre-training architecture for dense retrieval.
\newblock \emph{arXiv preprint arXiv:2104.08253}, 2021.

\bibitem[He et~al.(2022)He, Chen, Xie, Li, Doll{\'a}r, and Girshick]{mae}
Kaiming He, Xinlei Chen, Saining Xie, Yanghao Li, Piotr Doll{\'a}r, and Ross
  Girshick.
\newblock Masked autoencoders are scalable vision learners.
\newblock In \emph{CVPR}, 2022.

\bibitem[Hendrycks et~al.(2021{\natexlab{a}})Hendrycks, Basart, Mu, Kadavath,
  Wang, Dorundo, Desai, Zhu, Parajuli, Guo, Song, Steinhardt, and
  Gilmer]{rendition2021}
Dan Hendrycks, Steven Basart, Norman Mu, Saurav Kadavath, Frank Wang, Evan
  Dorundo, Rahul Desai, Tyler Zhu, Samyak Parajuli, Mike Guo, Dawn Song, Jacob
  Steinhardt, and Justin Gilmer.
\newblock The many faces of robustness: A critical analysis of
  out-of-distribution generalization.
\newblock In \emph{IEEE ICCV}, 2021{\natexlab{a}}.

\bibitem[Hendrycks et~al.(2021{\natexlab{b}})Hendrycks, Zhao, Basart,
  Steinhardt, and Song]{adversarial2021}
Dan Hendrycks, Kevin Zhao, Steven Basart, Jacob Steinhardt, and Dawn Song.
\newblock Natural adversarial examples.
\newblock In \emph{IEEE CVPR}, 2021{\natexlab{b}}.

\bibitem[Liu et~al.(2022)Liu, Hu, Lin, Yao, Xie, Wei, Ning, Cao, Zhang, Dong,
  Wei, and Guo]{swinv2}
Ze~Liu, Han Hu, Yutong Lin, Zhuliang Yao, Zhenda Xie, Yixuan Wei, Jia Ning, Yue
  Cao, Zheng Zhang, Li~Dong, Furu Wei, and Baining Guo.
\newblock Swin transformer v2: Scaling up capacity and resolution.
\newblock In \emph{International Conference on Computer Vision and Pattern
  Recognition (CVPR)}, 2022.

\bibitem[Radford et~al.(2021)Radford, Kim, Hallacy, Ramesh, Goh, Agarwal,
  Sastry, Askell, Mishkin, Clark, et~al.]{clip}
Alec Radford, Jong~Wook Kim, Chris Hallacy, Aditya Ramesh, Gabriel Goh,
  Sandhini Agarwal, Girish Sastry, Amanda Askell, Pamela Mishkin, Jack Clark,
  et~al.
\newblock Learning transferable visual models from natural language
  supervision.
\newblock In \emph{ICML}, pp.\  8748--8763. PMLR, 2021.

\bibitem[Ramesh et~al.(2021)Ramesh, Pavlov, Goh, Gray, Voss, Radford, Chen, and
  Sutskever]{dalle}
A.~Ramesh, Mikhail Pavlov, Gabriel Goh, Scott Gray, Chelsea Voss, Alec Radford,
  Mark Chen, and Ilya Sutskever.
\newblock Zero-shot text-to-image generation.
\newblock \emph{ArXiv}, abs/2102.12092, 2021.

\bibitem[Russakovsky et~al.(2015)Russakovsky, Deng, Su, Krause, Satheesh, Ma,
  Huang, Karpathy, Khosla, Bernstein, Berg, and Fei-Fei]{imagenet}
Olga Russakovsky, Jia Deng, Hao Su, Jonathan Krause, Sanjeev Satheesh, Sean Ma,
  Zhiheng Huang, Andrej Karpathy, Aditya Khosla, Michael Bernstein, Alexander~C
  Berg, and Li~Fei-Fei.
\newblock Imagenet large scale visual recognition challenge.
\newblock \emph{IJCV}, 2015.

\bibitem[van~den Oord et~al.(2017)van~den Oord, Vinyals, and
  Kavukcuoglu]{vqvae}
Aaron van~den Oord, Oriol Vinyals, and Koray Kavukcuoglu.
\newblock Neural discrete representation learning.
\newblock In \emph{NeurIPS}, NIPS'17, pp.\  6309–6318, Red Hook, NY, USA,
  2017. Curran Associates Inc.
\newblock ISBN 9781510860964.

\bibitem[Wang et~al.(2019)Wang, Ge, Lipton, and Xing]{sketch2019}
Haohan Wang, Songwei Ge, Zachary Lipton, and Eric~P Xing.
\newblock Learning robust global representations by penalizing local predictive
  power.
\newblock In \emph{Advances in Neural Information Processing Systems}, pp.\
  10506--10518, 2019.

\bibitem[Wang et~al.(2022)Wang, Bao, Dong, Bjorck, Peng, Liu, Aggarwal,
  Mohammed, Singhal, Som, and Wei]{beit3}
Wenhui Wang, Hangbo Bao, Li~Dong, Johan Bjorck, Zhiliang Peng, Qiang Liu, Kriti
  Aggarwal, Owais Mohammed, Saksham Singhal, Subhojit Som, and Furu Wei.
\newblock Image as a foreign language: {BEiT} pretraining for all vision and
  vision-language tasks.
\newblock \emph{ArXiv}, abs/2208.10442, 2022.

\bibitem[Wei et~al.(2021)Wei, Fan, Xie, Wu, Yuille, and
  Feichtenhofer]{maskfeat}
Chen Wei, Haoqi Fan, Saining Xie, Chao-Yuan Wu, Alan Yuille, and Christoph
  Feichtenhofer.
\newblock Masked feature prediction for self-supervised visual pre-training.
\newblock \emph{arXiv preprint arXiv:2112.09133}, 2021.

\bibitem[Wei et~al.(2022)Wei, Xie, Zhou, Li, and Tian]{mvp}
Longhui Wei, Lingxi Xie, Wengang Zhou, Houqiang Li, and Qi~Tian.
\newblock {MVP}: Multimodality-guided visual pre-training.
\newblock \emph{arXiv preprint arXiv:2203.05175}, 2022.

\bibitem[Xiao et~al.(2018)Xiao, Liu, Zhou, Jiang, and Sun]{upernet}
Tete Xiao, Yingcheng Liu, Bolei Zhou, Yuning Jiang, and Jian Sun.
\newblock Unified perceptual parsing for scene understanding.
\newblock In \emph{ECCV}, 2018.

\bibitem[Yu et~al.(2021)Yu, Li, Koh, Zhang, Pang, Qin, Ku, Xu, Baldridge, and
  Wu]{vitvqgan}
Jiahui Yu, Xin Li, Jing~Yu Koh, Han Zhang, Ruoming Pang, James Qin, Alexander
  Ku, Yuanzhong Xu, Jason Baldridge, and Yonghui Wu.
\newblock Vector-quantized image modeling with improved vqgan.
\newblock \emph{arXiv preprint arXiv:2110.04627}, 2021.

\bibitem[Zhai et~al.(2021)Zhai, Kolesnikov, Houlsby, and Beyer]{scaling:vit}
Xiaohua Zhai, Alexander Kolesnikov, Neil Houlsby, and Lucas Beyer.
\newblock Scaling vision {Transformers}.
\newblock \emph{arXiv preprint arXiv:2106.04560}, 2021.

\bibitem[Zhou et~al.(2019)Zhou, Zhao, Puig, Xiao, Fidler, Barriuso, and
  Torralba]{ade20k}
Bolei Zhou, Hang Zhao, Xavier Puig, Tete Xiao, Sanja Fidler, Adela Barriuso,
  and Antonio Torralba.
\newblock Semantic understanding of scenes through the {ADE20K} dataset.
\newblock \emph{Int. J. Comput. Vis.}, 127\penalty0 (3):\penalty0 302--321,
  2019.

\end{thebibliography}

\newpage
\appendix




\section{Visualization of Codebook}
\label{app:vis_code}

\begin{figure}[h]
\centering
\subfloat{
\includegraphics[width=0.98\textwidth]{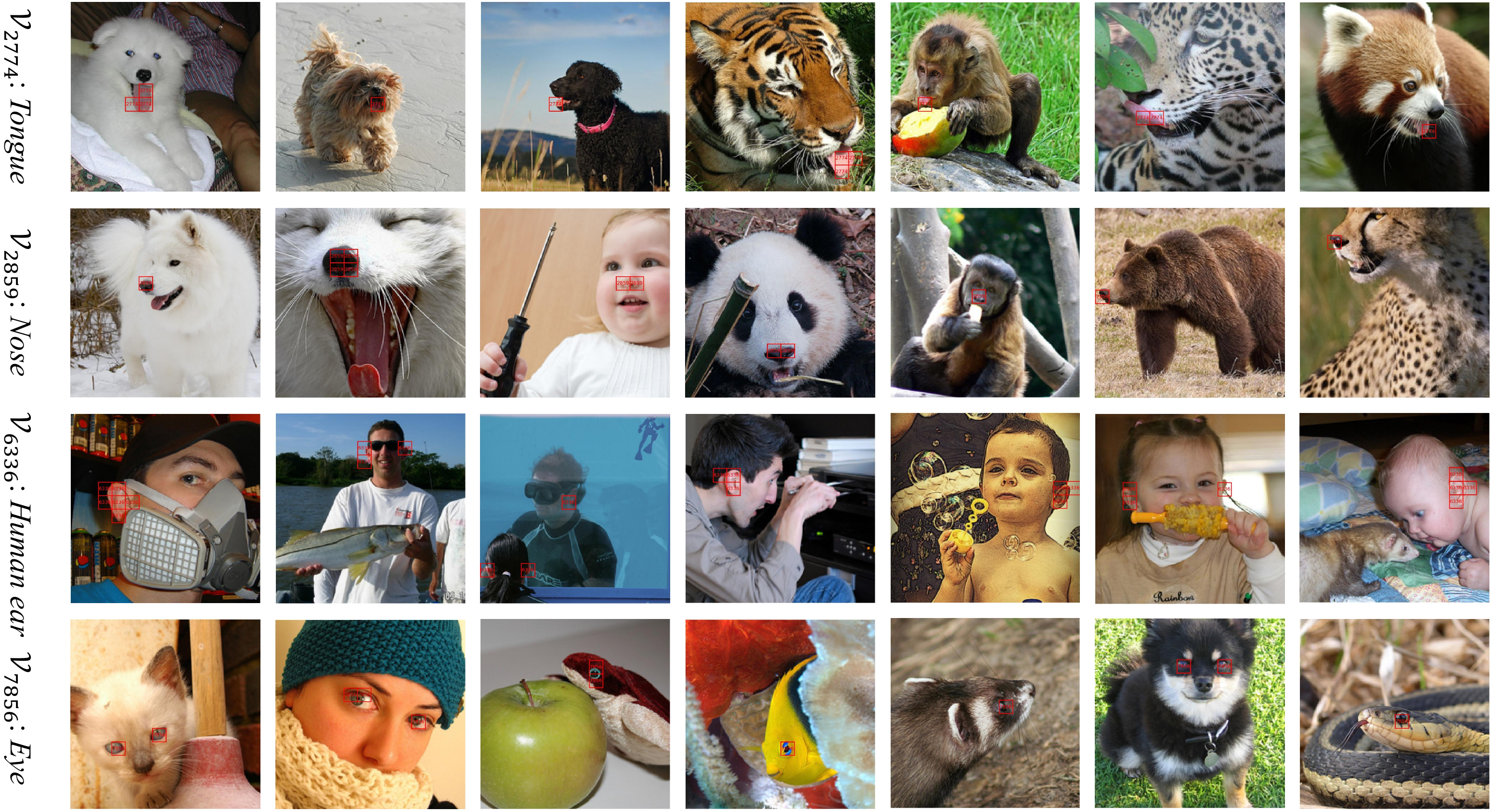}
}
\vspace{0.1in}
\subfloat{
\includegraphics[width=0.98\textwidth]{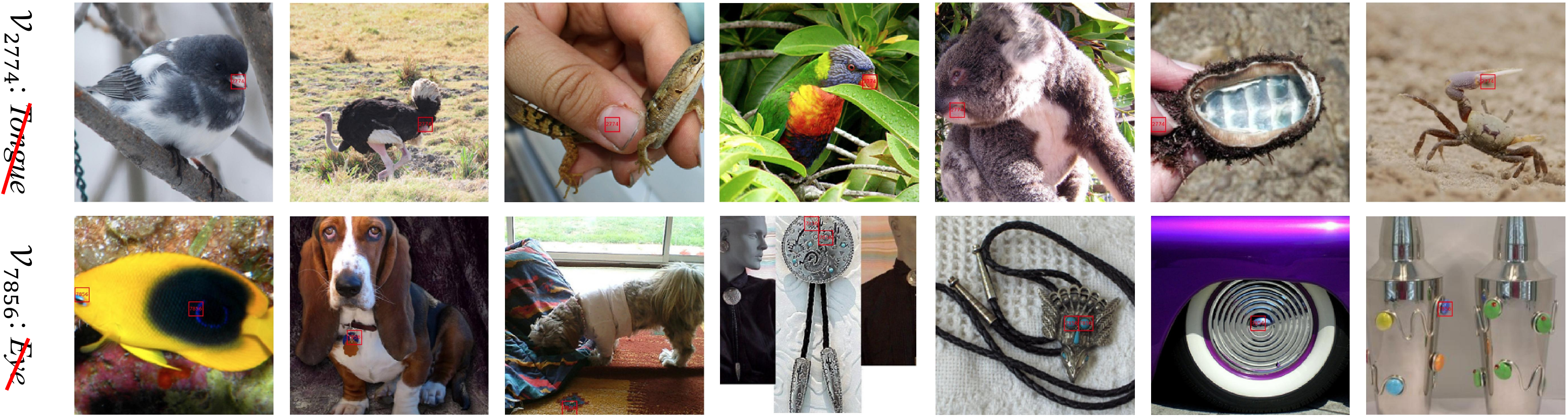}
}
\caption{Visualization of image patches corresponding to discrete codes. 
\textbf{Upper}: examples matching the learned semantic concepts; 
\textbf{Lower}: patches mis-matching the semantic concepts. Corresponding patches are marked in \fcolorbox{red}{white}{red rectangle}.
}
\label{fig:vis:code}
\end{figure}

It is observed that a discrete code tends to represent explicit semantics (Section~\ref{sec:analysis}). In Figure~\ref{fig:vis:code}(upper), we show image examples corresponding to a given discrete code. One can see that discrete codes ignore image details, such as color, illumination, rotation and scale. 

In the lower part of Figure~\ref{fig:vis:code}, we also show some patches that mismatch the semantic concepts.
Taking the fish (the first image at the last row) as instance, \vqkd{} misclassifies the spot on the fish body as the eye concept due to the local structure similarity.

\section{Comparison with Large-scale Supervised Pretraining}
\label{app:large_scale}

We report the performance by using the ImageNet-1k for pretraining in Table~\ref{tbl:results}. To show the data scalability of \our{}, we conduct intermediate fine-tuning experiments on ImagNet-21k and final fine-tuning on ImageNet-1k, by using the 1600 epoch pretraining models in Table~\ref{tbl:results}. From Table~\ref{tbl:supervised:pt:cls}, \our{} using ViT-L/16 with $384\times384$ input resolution, achieves 89.0\% top-1 accuracy, which even outperforms ViT-H/14 using Google JFT-3B labeled dataset by 0.5\%.
This significant performance gain indicates the data efficiency and superiority of the proposed \our{}.

\begin{table}[h]
\centering
\caption{Top-1 accuracy on ImageNet-1K fine-tuning. $224^2$ and $384^2$ denote model resolutions.
}
\label{tbl:supervised:pt:cls}
\begin{tabular}{lcccc}
\toprule
\multirow{2}{*}{\bf Models}  &  \multirow{2}{*}{\bf Model Size}& \multirow{2}{*}{\bf Labeled Data Size} & \multicolumn{2}{c}{\bf ImageNet-1k} \\
 &  &   & $\textbf{224}^\textbf{2}$ &  $\textbf{384}^\textbf{2}$ \\
\midrule
\multicolumn{5}{l}{\textit{Supervised Pretraining on ImageNet-21K}} \\
ViT-B/16~\citep{vit} & 86M & 14M & - & 84.0  \\
ViT-L/16~\citep{vit} & 307M & 14M & - & 85.2 \\
ViT-H/14~\citep{vit} & 632M & 14M & - & 85.1 \\
\midrule
\multicolumn{5}{l}{\textit{Supervised Pretraining on Google JFT-300M (using labeled data)}} \\
ViT-B/16~\citep{vit} & 86M & 300M & - & 84.2  \\
ViT-L/16~\citep{vit} & 307M & 300M & - & 87.1 \\
ViT-H/14~\citep{vit} & 632M & 300M & - & 88.0 \\
\midrule
\multicolumn{5}{l}{\textit{Supervised Pretraining on Google JFT-3B }} \\
ViT-B/16~\citep{scaling:vit} & 86M & 3000M & - & 86.6 \\
ViT-L/16~\citep{scaling:vit} & 307M & 3000M & - & 88.5 \\
\midrule
\multicolumn{5}{l}{\textit{\beit{} Pretraining on ImageNet-21K, and Intermediate Fine-Tuning on ImageNet-21K}} \\
\beit{} ViT-B/16~\citep{beit} & 86M & 14M & 85.2  & 86.8 \\
\beit{} ViT-L/16~\citep{beit} & 307M & 14M & 87.4 & 88.4 \\
\midrule
\multicolumn{5}{l}{\textit{\our{} Pretraining on ImageNet-1K, and Intermediate Fine-Tuning on ImageNet-21K}} \\
\our{} ViT-B/16 (ours) & 86M & 14M & 86.5 & 87.5 \\
\our{} ViT-L/16 (ours) & 307M & 14M & \bf 88.4 & \bf 89.0 \\
\bottomrule
\end{tabular}
\end{table}

\section{Overall Framework for \our{}}
\label{app:overall_framework}

\begin{figure}[H]
\begin{center}
\begin{tabular}{c}
\includegraphics[width=1\textwidth]{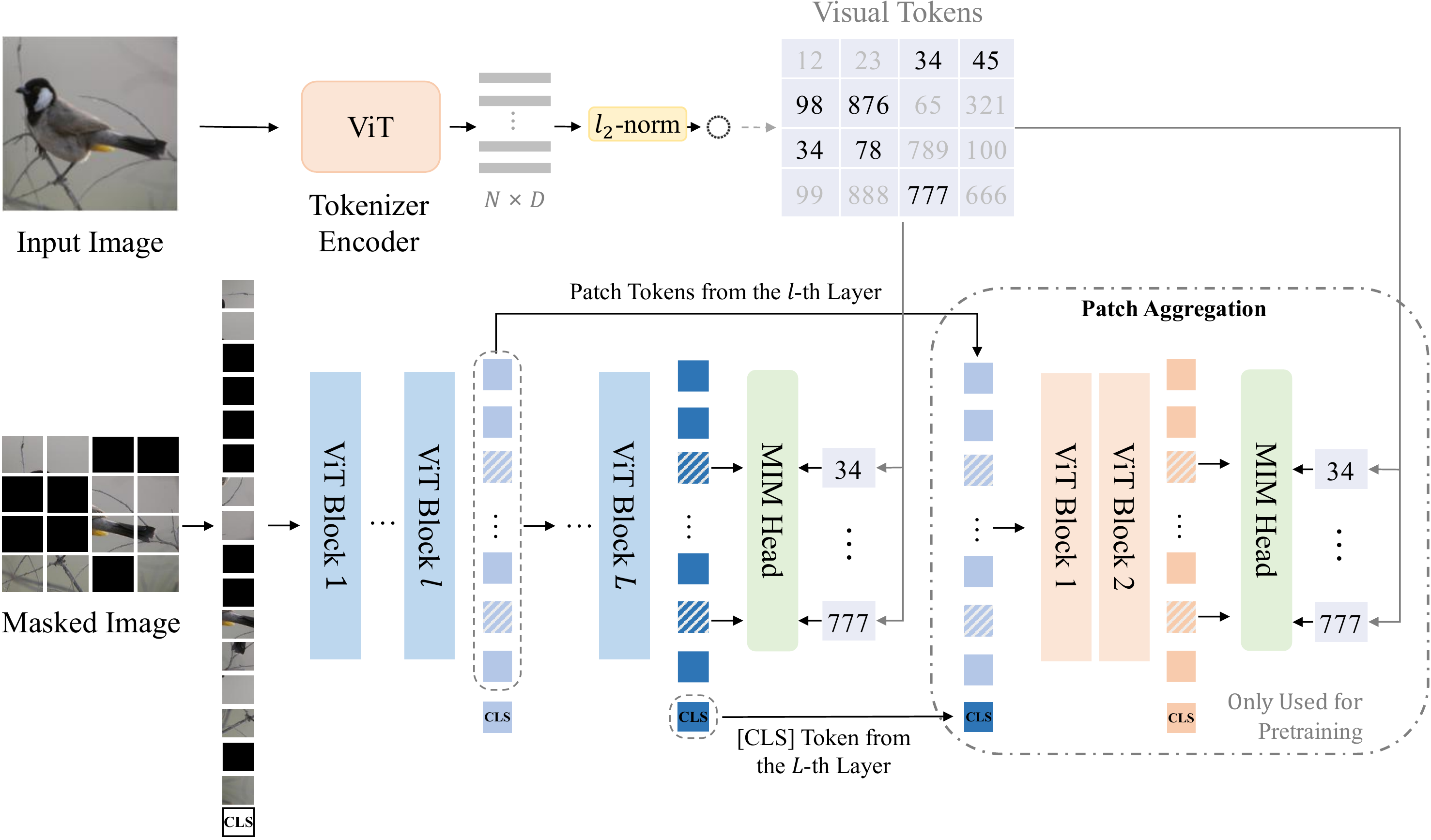}
\end{tabular}
\end{center}
\caption{Overall framework for \our{} pretraining.
}
\label{fig:v2_overall_framework}
\end{figure}

We show the tokenizer training part and \our{} pretraining part in Figure~\ref{fig:vqkd} and Figure~\ref{fig:beitv2}, respectively.
In addition, we present the whole pretraining process in Figure~\ref{fig:v2_overall_framework}.

\section{Hyperparameters for \vqkd{} Training}
\label{app:vqkd}

\begin{table}[H]
\centering
\caption{
Hyperparameters for training \vqkd{} on ImageNet-1K.
}
\label{tbl:vqkd_param}
\scalebox{0.98}{
\begin{tabular}{l|c}     
\toprule
\bf Hyperparameters &  \bf Values \\
\midrule
Encoder layers & 12 \\
Decoder layers & \{1, 3\} \\
Hidden size & 768 \\
FFN inner hidden size & 3072  \\
Attention heads & 12  \\
Attention head size & 64 \\
Patch size & $16 \times 16$ \\
Codebook size & $8192 \times 32$ \\
\midrule
Training epochs & 100 \\
Batch size & 512 \\
Adam $\beta$ & (0.9, 0.99) \\
Peak learning rate & 2e-4 \\
Minimal learning rate & 1e-5 \\
Learning rate schedule & Cosine \\
Warmup epochs & 5 \\
\midrule
Gradient clipping & \xmark \\
Dropout & \xmark \\
Stoch. depth & \xmark \\
Weight decay & 1e-4 \\
\midrule
Data Augment & RandomResizeAndCrop \\
Input resolution & $224 \times 224$ \\
\bottomrule
\end{tabular}
}
\end{table}

\section{Hyperparameters for \our{} pretraining}
\label{app:pretrain}

\begin{table}[H]
\centering
\small
\caption{
Hyperparameters for \our{} pretraining on ImageNet-1K. $^*$ denotes that the hyperparameters are adopted when the pretraining schedule is 300 epochs.
}
\label{tbl:pretrain:hyperparams}
\scalebox{0.98}{
\begin{tabular}{l|cc}
\toprule
\bf Hyperparameters & \bf Base Size & \bf Large Size \\
\midrule
Layers & 12 & 24 \\
Hidden size & 768 & 1024 \\
FFN inner hidden size & 3072 & 4096 \\
Attention heads & 12 & 16 \\
Layer scale & 0.1 & 1e-5 \\
Patch size & \multicolumn{2}{c}{$16 \times 16$} \\
Relative positional embeddings & \multicolumn{2}{c}{\cmark} \\
Shared relative positional embeddings & \multicolumn{2}{c}{\cmark} \\
\midrule
Training epochs & \multicolumn{2}{c}{300$^*$/1600} \\
Batch size & \multicolumn{2}{c}{2048} \\
Adam $\beta$ & \multicolumn{2}{c}{(0.9, 0.98$^*$/0.999)} \\
Peak learning rate & \multicolumn{2}{c}{1.5e-3} \\
Minimal learning rate & \multicolumn{2}{c}{1e-5} \\
Learning rate schedule & \multicolumn{2}{c}{Cosine} \\
Warmup epochs & \multicolumn{2}{c}{10} \\
\midrule
Gradient clipping & \multicolumn{2}{c}{3.0} \\
Dropout & \multicolumn{2}{c}{\xmark} \\
Drop path & \multicolumn{2}{c}{0$^*$/0.1} \\
Weight decay & \multicolumn{2}{c}{0.05} \\
\midrule
Data Augment & \multicolumn{2}{c}{RandomResizeAndCrop} \\
Input resolution & \multicolumn{2}{c}{$224 \times 224$} \\
Color jitter & \multicolumn{2}{c}{0.4} \\
\bottomrule
\end{tabular}
}
\end{table}

\section{Hyperparameters for Image Classification Fine-tuning}
\label{app:finetune:cls}

\begin{table}[H]
\centering
\caption{
Hyperparameters for fine-tuning \our{} on ImageNet-1K.
}
\label{tbl:ft:imagenet:hyperparams}
\scalebox{0.95}{
\begin{tabular}{l|cc}
\toprule
\bf Hyperparameters & \bf ViT-B/16 & \bf ViT-L/16 \\
\midrule
Peak learning rate & 5e-4 & 5e-4\\
Fine-tuning epochs & 100  & 50 \\
Warmup epochs & 20 & 5 \\
Layer-wise learning rate decay & 0.65 & 0.8 \\
Batch size & \multicolumn{2}{c}{1024} \\
Adam $\epsilon$ & \multicolumn{2}{c}{1e-8}  \\
Adam $\beta$ & \multicolumn{2}{c}{(0.9, 0.999)} \\
Minimal learning rate & \multicolumn{2}{c}{1e-6} \\
Learning rate schedule & \multicolumn{2}{c}{Cosine} \\
\midrule
Repeated Aug & \multicolumn{2}{c}{\xmark} \\
Weight decay & \multicolumn{2}{c}{0.05} \\
Label smoothing $\varepsilon$ & \multicolumn{2}{c}{0.1}     \\
Stoch. depth & 0.1 & 0.2 \\
Dropout & \multicolumn{2}{c}{\xmark} \\
Gradient clipping & \multicolumn{2}{c}{\xmark} \\
\midrule
Erasing prob.  & \multicolumn{2}{c}{0.25} \\
Input resolution & \multicolumn{2}{c}{$224 \times 224$} \\
Rand Augment  & \multicolumn{2}{c}{9/0.5} \\
Mixup prob.  & \multicolumn{2}{c}{0.8}     \\
Cutmix prob.   & \multicolumn{2}{c}{1.0}    \\
\midrule
Relative positional embeddings & \multicolumn{2}{c}{\cmark} \\
Shared relative positional embeddings & \multicolumn{2}{c}{\xmark} \\
\bottomrule
\end{tabular}
}
\end{table}

\section{Hyperparameters for ADE20K Semantic Segmentation Fine-tuning}
\label{app:finetune:seg}

\begin{table}[H]
\centering
\caption{
Hyperparameters for fine-tuning \our{} on ADE20K.
}
\label{tbl:ft:ade20k:hyperparams}
\begin{tabular}{l|c c}
\toprule
\bf Hyperparameters & \bf ViT-B/16 & \bf ViT-L/16 \\
\midrule
Input resolution & \multicolumn{2}{c}{$512 \times 512$} \\
\midrule
Peak learning rate & \multicolumn{2}{c}{\{0.5, 0.8, 1.0\}e-4} \\
Fine-tuning steps & \multicolumn{2}{c}{160K} \\
Batch size & \multicolumn{2}{c}{16} \\
Adam $\epsilon$ & \multicolumn{2}{c}{1e-8}  \\
Adam $\beta$ & \multicolumn{2}{c}{(0.9, 0.999)} \\
Layer-wise learning rate decay & \multicolumn{2}{c}{\{0.75, 0.8, 0.85\}} \\
Minimal learning rate & \multicolumn{2}{c}{0} \\
Learning rate schedule & \multicolumn{2}{c}{Linear} \\
Warmup steps & \multicolumn{2}{c}{1500} \\
\midrule
Dropout & \multicolumn{2}{c}{\xmark} \\
Stoch. depth & 0.1 & 0.2 \\
Weight decay & \multicolumn{2}{c}{0.05} \\
\midrule
Relative positional embeddings & \multicolumn{2}{c}{\cmark} \\
Shared relative positional embeddings & \multicolumn{2}{c}{\xmark} \\
\bottomrule
\end{tabular}
\end{table}

\end{document}